\documentclass[conference]{IEEEtran}
\IEEEoverridecommandlockouts
\usepackage{cite}
\usepackage{amsmath,amssymb,amsfonts}
\usepackage{algorithmic}
\usepackage{graphicx}
\usepackage{caption}
\usepackage{subcaption}
\usepackage{textcomp}
\usepackage{xcolor}
\def\BibTeX{{\rm B\kern-.05em{\sc i\kern-.025em b}\kern-.08em
    T\kern-.1667em\lower.7ex\hbox{E}\kern-.125emX}}

\usepackage{url}
\usepackage{hyperref}

\ifCLASSOPTIONcompsoc
  \usepackage[nocompress]{cite}
\else
  \usepackage{cite}
\fi

\usepackage{natbib}

\begin{document}

\title{DIVERSE: A Dataset of YouTube Video Comment Stances with a Data Programming Model
}

\author{\IEEEauthorblockN{Anonymous Authors}}

\author{\IEEEauthorblockN{Iain J. Cruickshank}
\IEEEauthorblockA{\textit{Army Cyber Institute} \\
Highland Falls, New York, USA \\
iain.cruickshank@westpoint.edu}
\and
\IEEEauthorblockN{Lynnette Hui Xian Ng}
\IEEEauthorblockA{\textit{Carnegie Mellon University} \\
Pittsburgh, Pennsylvania, USA \\
lynnetteng@cmu.edu}
\and
\IEEEauthorblockN{Amir Soofi}
\IEEEauthorblockA{\textit{University of California, Los Angeles} \\
Los Angeles, California, USA \\
soofi@cs.ucla.edu}
}

\maketitle

\begin{abstract}
Public opinion of military organizations significantly influences their ability to recruit talented individuals. As recruitment efforts increasingly extend into digital spaces like social media, it becomes essential to assess the stance of social media users toward online military content. However, there is a notable lack of data for analyzing opinions on military recruiting efforts online, compounded by challenges in stance labeling, which is crucial for understanding public perceptions. Despite the importance of stance analysis for successful online military recruitment, creating human-annotated, in-domain stance labels is resource-intensive. In this paper, we address both the challenges of stance labeling and the scarcity of data on public opinions of online military recruitment by introducing and releasing the DIVERSE dataset: \footnote{\url{https://doi.org/10.5281/zenodo.10493803}}. This dataset comprises all comments from the U.S. Army's official YouTube Channel videos. We employed a state-of-the-art weak supervision approach, leveraging large language models to label the stance of each comment toward its respective video and the U.S. Army. Our findings indicate that the U.S. Army’s videos began attracting a significant number of comments post-2021, with the stance distribution generally balanced among supportive, oppositional, and neutral comments, with a slight skew towards oppositional versus supportive comments.
\end{abstract}

\begin{IEEEkeywords}
stance detection, social media, large language models, ensemble learning
\end{IEEEkeywords}

\section{Introduction}
The U.S. military is facing a recruitment crisis. For the last few years, military organizations like the U.S. Army have struggled to meet recruiting targets to maintain their desired force structure \citep{Brown:2023}. Social media plays an important role in discerning stances \citep{aldayel2021stance}, and in recruitment \citep{ng2023recruitment}. Thus, it is critical to understand how messages from the military are received on social media, as it can shape recruiting messaging. More broadly speaking, the ability to evaluate how social media content is being received by a target audience is crucial for any marketing campaign, be it a war recruiting effort or a organizational recruitment effort or for influencing consumer behavior.

A particular indicator of the reception of social media content is stance. Stance detection, or the determination of an expressed or implied opinion toward a target, remains a primary task for many computational social studies. Several articles have proposed methods and datasets for the task of stance detection \citep{allaway2023zero,du2007stance}. However, existing datasets explore stance as expressed on a single social media platform, mainly Twitter/X, and only have a single stance per comment \citep{sobhani2017dataset,wtwt, mohammad2016semeval,covidlies,phemerumours,villa2020stance}. Since stance detection remains a difficult task due to the context-dependent nature of the definition of stance (e.g., the requirement of a target for stance to be defined), along with varying definitions of stance \citep{ng2022my}, all of the aforementioned stance datasets rely on human annotation to create the stance labels. This can result in human errors being present in the datasets as well as greatly reduce the size of the datasets to be in the thousands since human annotation requires a lot of resources and time.

To address these shortcomings in prior stance datasets, we propose a new stance dataset based on YouTube comments and labeled with machine-assisted labeling through the data programming paradigm of weak supervision. This dataset is a collection of the YouTube comments posted on videos uploaded by the U.S. Army\footnote{Offical U.S. Army YouTube Channel: \url{https://www.youtube.com/@usarmy}}. The comments are first annotated for weak labels representing aspects of stance expression (hate speech, sarcasm, sentiment). The comments are then labeled by prompting three different LLMs. Finally, these weak labels are combined to produce the final stance labels of the comments towards the U.S. Army and its videos. With this data set, we can better understand critical issues with the public opinion toward the U.S. Army and its recruitment efforts.

Our contributions in this paper are as follows:
\begin{itemize}
    \item \textbf{A novel benchmark stance dataset of YouTube comments.} We present a dataset for the task of stance detection that has several novel differences from previous benchmark datasets. While most stance datasets use posts from Twitter, ours provides posts from the YouTube platform instead, contributing to the breadth of datasets. Another difference is that our dataset labels stances towards more than one target in a comment (the only other being \citet{sobhani2017dataset}), and provides labels to weak signals including the presence of contentious issues, conspiracy theories, and other emotional content. Finally, this dataset has more than three times the number of labeled comments over the current largest benchmark dataset (\citet{wtwt} at $~32,000$ data points vs $>$216,000 data points in ours).
    \item \textbf{The use of a data-programming based stance-labeling methodology.} We combine recent advances for creating large, high-quality labeled datasets with contextual knowledge to generate quality stance labels \cite{ratner2019training, huang2024alchemist, smith2024language}. This involves making use of weak labels such as the presence of hate speech, the sentiment of the content, the presence of sarcasm, known keywords, and the evaluation of Large Language Models (LLMs), which are combined using weak supervision to determine the final stance label.
\end{itemize}

\section{Related Work}
\paragraph*{Stance Datasets} The task of stance detection has led to the development of several datasets. \citet{ng2022my} consolidated several stance detection datasets with social media posts across multiple topics. These datasets use three basic labels -- favor, against, neither -- to express how the sentence views a target \citep{mohammad2016semeval}. There are datasets that use labels of implicit and explicit support and denial to indicate the overtness of stance expression \citep{qazvinian2011rumor}. Automated stance detection algorithms have been constructed with these datasets using support vector machines \citep{elfardy2016cu}, logistic regressions \citep{augenstein2016usfd}, neural network models \citep{fang2019neural,siddiqua2019tweet} and deep learning models \citep{kuccuk2020stance}. Recently, Large-Language Models have also been used for identifying stance in sentences \citep{zhang2023logically,gatto2023chain}.

\paragraph*{YouTube Datasets} To date, there have been several research investigations involving YouTube, but few datasets. For example, there are a number of papers investigating influence campaigns \citep{marcoux2021analyzing, hussain2018analyzing}, conspiracy theories \citep{liaw2023younicon}, comment toxicity \citep{obadimu2019identifying}, and even cross-platform use of YouTube videos \citep{ginossar2022cross}. Despite the focus on analyzing YouTube data, including comment data, most research projects do not release their datasets \citep{liaw2023younicon}. Furthermore, to the best of the authors' knowledge, there are no datasets or investigations of YouTube data for stance detection, particularly concerning the comments section on YouTube videos. It is important to note, however, that there are a number of commercially available social media suites to analyze comments to posts on platforms for indicators like sentiment and influencer engagement, but none of these tools provide stance detection \citep{ng2022my, rogan:2022}.

\paragraph*{Large Language Models for Stance Detection} Recently, researchers have explored the application of Large Language Models (LLMs) in stance detection. Existing studies primarily focused on ChatGPT, yielding mixed outcomes. \citet{zhang2022would} demonstrated improved results on the SemEval2016 benchmark dataset using ChatGPT with instruction-based prompts. Other works have also found good, if not necessarily state-of-the-art performance, on the task of stance detection for LLMs \citep{ziems2023can,liyanage2023gpt, zhang2023investigating}. Conversely, \citet{aiyappa2023can} raised concerns about potential data contamination from ChatGPT's extensive training dataset, casting doubt on the reliability of evaluations. \citet{mets2023automated} examined ChatGPT's usability as a zero-shot classifier for stance detection, reporting competitive but ultimately inferior performance compared to supervised models. Despite these findings, the effectiveness of LLMs in stance classification remains promising, if uncertain, especially without fine-tuning on labeled data or prompt engineering.

\paragraph*{Weak Supervision for Dataset creation} Weak Supervision has emerged as a promising approach for expeditiously generating large training datasets. This methodology leverages heuristic rules and weak supervision to programmatically label data at scale \citep{ratner2016data, ratner2019training}. Rather than relying solely on manually annotated datasets, weak supervision enables the automatic creation of training sets by combining various weak sources, derived from heuristics, distant supervision, and domain expertise. These weaker, noisy labels can then be combined to produce high-quality labels. This paradigm shift offers an efficient means of tackling data scarcity challenges, especially in scenarios where obtaining labeled data is time-consuming or expensive. The incorporation of weak supervision into the realm of stance detection holds the potential to create quality labeled datasets with much less human annotation.

Finally, it is important to note that some parallel works to ours have explored combining weak supervision with LLMs to create labels. In \cite{smith2024language}, the authors prompt an LLM to provide possible labels for examples and then combine these weak labels through weak supervision to produce final labels, which are significantly better than what an LLM can produce via zero-shot. \cite{huang2024alchemist} use an LLM to create weak labeling functions for a dataset and then label the dataset with those weak labeling functions and, finally, combine the weak labels through weak supervision to produce a labeled dataset. In all cases, the combination of an LLM and weak supervision produces better improved labeling outcomes from a zero-shot scenario.

\section{Dataset Creation}
This dataset was collected to understand the reception of YouTube users toward the US Army's marketing material on YouTube, thereby understanding their support/opposition towards the military. Each comment is labeled for two stances: (a) the stance of the comment towards its source video, and (b) the stance of the comment towards the US Army. There are three possible options for stances: ``supports", ``against", and ``neutral". The following parts of this section details the data collection and labeling procedures.

\subsection{Data Collection}
 We collected comments on the videos of the US Army's official YouTube channel (username: @usarmy\footnote{\url{https://www.youtube.com/usarmy}}) using the YouTube Data API\footnote{\url{https://developers.google.com/youtube/v3/docs/comments/list}}. This returned both comments as well as replies to comments. Collection took place from 2 to 5 October 2023 and again from 2-9 April 2024, where we retrieved a total of 216,385 comments across 1,082 videos. For each comment, we collected the following information:

\begin{itemize}
    \item \textbf{id}: the unique id of a comment
    \item \textbf{comment}: the text of the comment
    \item \textbf{author}: a unique id for the author of the comment
    \item \textbf{like\_count}: the likes the comment received
    \item \textbf{published\_at}: the time and date of the comment's publication
    \item \textbf{conversation\_id}: a unique id of the conversation that a comment belongs to. This is also the id of the root comment
    \item \textbf{video\_id}: the unique id of the video of the comment 
    \item \textbf{name}: the title of the video of the comment 
    \item \textbf{description}: a text description of the video of the comment
    \item \textbf{timestamp}: the time and date of the publication of the video the comment is commenting on
\end{itemize}

\subsection{Data Annotation Methodology}
\begin{figure*}[!ht]
    \centering
    \includegraphics[width=\textwidth]{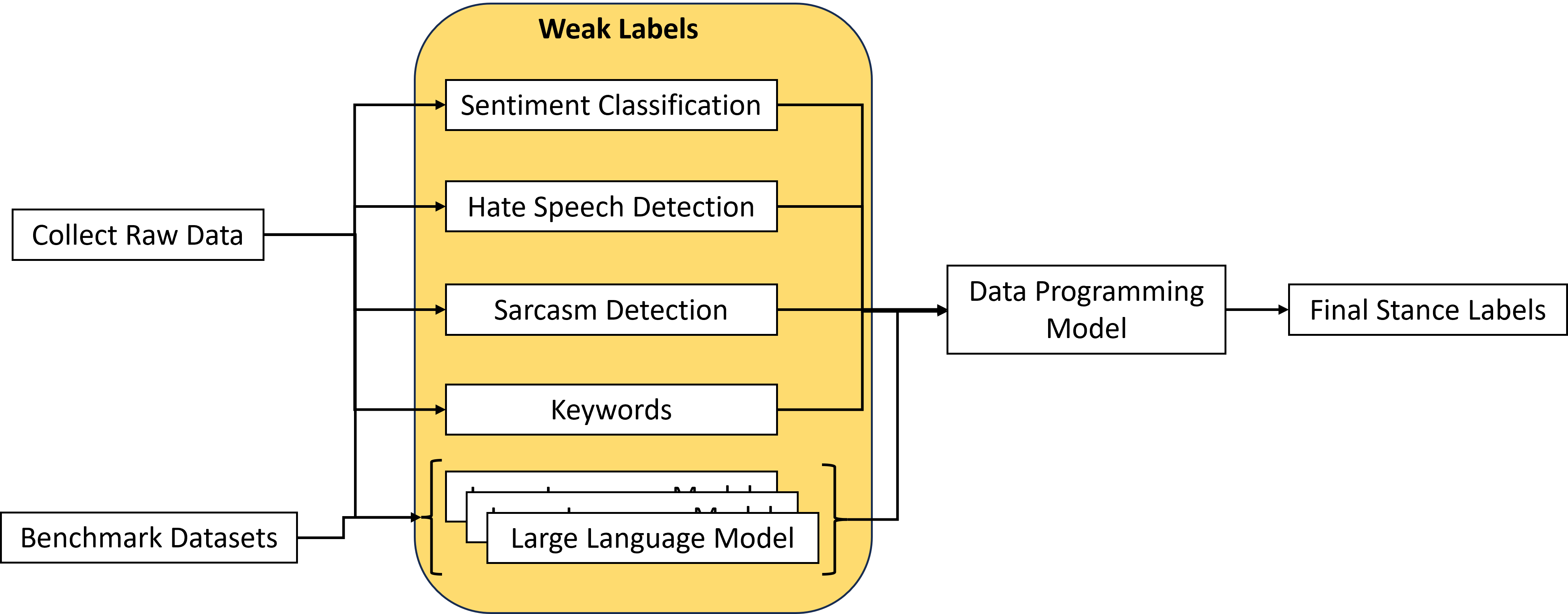}
    \caption{Pipeline of stance labeling methodology}
    \label{fig:dataset_creation_and_labeling}
\end{figure*}

Instead of using human annotators for stance labeling, we used weak supervision \cite{ratner2016data, ratner2019training}. This paradigm produces a final, more accurate label using a series of lesser quality labels, \textit{weak labels}, which may capture different aspects of the concept being labeled. Such labeling techniques have been employed in labeling happiness with cognitive appraisal dimension \cite{liu2023psyam} and negotiation strategies using the presence of reasoning and friendliness \cite{jaidka2023takes}. We used a combination of \textit{weak labels} based upon subject matter expertise, domain expertise, machine learning models, and recent advances in the use of LLMs for stance detection. These labels were then combined through a weak supervision model to produce the final labels. In this way, we include both specialty functions similar to \cite{huang2024alchemist} as well as LLM-direct labels in \cite{smith2024language} into the weak supervision process. \autoref{fig:dataset_creation_and_labeling}, depicts the entire data annotation methodology workflow. We detail key aspects of the workflow in the following sections. 

\subsection{Annotating Weak Labels}
To create the weak labels we leveraged dataset-specific traits, domain expertise and LLMs. For example, in this dataset, the presence of hate speech or sarcasm frequently indicates the comment being against the video and by extension, the user that uploads it, which is the US Army. Therefore, the presence of these weak signals can be combined to indicate the final label. It is important to note for any given weak label, it need not produce a label for every comment in the dataset. Some of the weak labels simply do not apply to some of the comments, in which case they have an ``abstain" response, which is distinct from the possible stance labels, and are not used in the computation of the final stance labels.

\subsubsection{Hate Speech} Hate Speech is defined as speech to incite violence or hate, and typically has specific targets like ethnic groups or religion \cite{10.1145/3232676}. In the case of this dataset and stance, we noticed that many of the comments against both the posted videos and the US Army more generally featured aspects of hate speech, like antisemitic and misogynistic content. Several machine learning algorithms ranging from logistic regression to neural network models have been constructed to perform automatic hate speech detection \cite{10.1145/3232676}. For this weak label, we used the hate speech classification model from \cite{kralj2022handling} \footnote{\url{https://huggingface.co/IMSyPP/hate_speech_en}}. This model was used as is without any fine-tuning. For each input text, this model returns three different types of ``hate" label or an ``acceptable" label. We considered the ``hate" response as an ``against" stance label, and any other response as an ``abstain."

\subsubsection{Sarcasm} Sarcasm is a literary device to convey one's intention in an implicit manner \cite{chaudhari2017literature}. As with hate speech, we found that commenters on these YouTube videos often use sarcasm as a way of attacking the video or the entity behind the video. We used a multilingual sarcasm detector\footnote{\url{helinivan/multilingual-sarcasm-detector}} which trains a BERT-based sarcasm detector from newspapers text in English, Dutch and Italian. For each input text, this model returns a label of ``sarcasm" or ``no\_sarcasm", and we considered the ``sarcasm" response as an ``against" stance label, and any other response as an ``abstain."

\subsubsection{Sentiment} Sentiment analysis provides an understanding of the positivity or negativity of public sentiment towards a topic \cite{10.1145/3604605}. In the context of stance detection and classification, while stance is distinct from sentiment, the two often overlap; a negative sentiment often indicates a stance against a target and vice-versa for a positive sentiment. We used a multilingual-cased sentiment analysis model\footnote{\url{lxyuan/distilbert-base-multilingual-cased-sentiments-student}} that performs sentiment-analysis using a teacher-student architecture with the DistillBERT model. For each input text, this model returns a score between 0 and 1 for three possible sentiment labels of ``positive", ``negative", and ``neutral". We considered a ``positive" sentiment greater than 0.75 as a ``supports" stance and a ``negative" sentiment greater than 0.75 as a ``against" stance and any other value as a ``abstain" stance.

\subsubsection{Use of Keywords} We manually determined a set of keywords and phrases that when present, would represent either one of the non-neutral stance labels. For each comment, we identified if any words from the set of keywords were present and assigned the corresponding stance to this weak label. For example, keywords and phrases like ``hooah", ``awesome", and ``god bless" were associated with a ``supports" stance, while ``woke", ``murder", and ``propaganda" were associated with an ``against" stance. the full list of these words is available in our released code\footnote{\url{https://doi.org/10.5281/zenodo.10493803}}.

\subsubsection{Stance from Large Language Models} Based on the recent work demonstrating effectiveness of LLMs in the stance classification task \citep{ng2024examining}, we also directly labeled stances using LLMs.

\paragraph*{Pre-tuning of LLMs}
Due to the size of the dataset, we opted to use open-source LLMs which could be run on local hardware. Specifically, we used three different LLMs: (a) Flan-UL2 \citep{tay2022unifying}, which is an encoder-decoder architecture, (b) Mistral-7B \citep{jiang2023mistral}, which is a decoder-only architecture, and (c) Llama-3-8B \cite{dubey2024llama}, which is another decoder-only architecture. We used the Flan-UL2 and Llama-3 models in a zero-shot setting only (i.e., we did not perform any fine-tuning on the model) \citep{ziems2023can}. For the Mistral-7B model, we both used it in a zero-shot setting and we fine-tuned the model on a collection of benchmark stance datasets, including Semeval2016 \citep{mohammad2016semeval}, Election2016 \citep{sobhani2017dataset}, srq \citep{villa2020stance}, wtwt \citep{wtwt}, phemerumors \citep{phemerumours}, and covid-lies \citep{covidlies}. Based on the research in \citep{ng2022my}, which found that the inclusion of more benchmark datasets into training improved the generalizability of the trained model, hence we trained on all of the datasets. For the training of the Mistral LLM, we used a LoRA adapter \citep{hu2021lora} with parameters of $\alpha=1$, $r=16$, $dropout=0.1$, and no bias term. The model was trained for two epochs using an AdamW optimizer \citep{loshchilov2017decoupled}, a learning rate of $1e-4$, and a linear learning rate scheduler with 50 warm-up steps. To feed the data to the model during training, we adopted a prompt that included context about stance classification, as has been done in zero-shot prompting for stance in other works. The following was the training prompt:

\begin{quote}
```Stance classification is the task of determining the expressed or implied opinion, or stance, of a statement toward a certain, specified target. Analyze the following social media statement and determine its stance towards the provided [dataset entity]. 

Respond with a single word: ``for", ``against",``neutral".

[dataset entity]: \{entity\}

statement: \{statement to classify\}

stance: '''
\end{quote}

Where the [dataset entity] was varied depending on the benchmark dataset (e.g., for Election2016, it was `politician', while for phemerumors it was `rumor'). For the srq dataset, we changed the prompt slightly to incorporate the post that the reply is replying to as the task in that dataset is to label the stance of the reply to the post, versus the stance of the post about a topic or person. The loss was determined by whether the model could output the tokens for the correct stance, and only those tokens. The training of the model was done on a local computer running Ubuntu Linux with a 96-core processor and three A6000 GPUs. Overall, this resulted in four, different LLMs for labeling the comments: one off-the-shelf UL2 model, one off-the-shelf Llama-3 model and two Mistral models.

Finally, we also fine-tuned a small BERT model as was done in \cite{ng2022my} on all of the benchmark datasets previously described and labeled the stances with that fine-tuned model as well.

\paragraph*{Analyzing stances with LLMs} In analyzing the comments for stances using LLMs, we used two different prompts, one that only included the comment, and one that included the comment that a comment was replying to, if it was replying to a comment. The labeling prompt without incorporating the reply structure is:

\begin{quote}
```Analyze the following YouTube comment to a video posted by the U.S. Army named ``\{title\}" and determine its stance towards the provided entity. Respond with a single word: ``for", ``against", ``neutral", or ``unrelated". Only return the stance as a single word, and no other text.

entity: \{entity\}

comment: \{comment to classify\}

stance: 

'''
\end{quote}

while the labeling prompt that did incorporate the reply structure is:

\begin{quote}
```Analyze the following YouTube comment to a video posted by the U.S. Army named ``\{title\}" and determine its stance towards the provided entity. If the statement is a reply to another YouTube comment, that YouTube comment is listed below. Respond with a single word: ``for", ``against", ``neutral", or ``unrelated". Only return the stance as a single word, and no other text.

entity: \{entity\}

previous comment: \{previous\_comment\}

comment: \{comment to classify\}

stance: 

'''
\end{quote}

We also experimented with a question answering phrasing of the task during labeling, but found the analyze phrasing to generally perform slightly better in benchmark datasets. Overall, each comment was labeled for a stance by an LLM seven different times: once with the reply prompt and once without the reply prompt for the UL2 model, the Llama-3 model, and other Mistral model, and with the without the reply prompt by the stance-tuned Mistral model. Since we are labeling the comment for its stance toward the US Army as well as the video it is commenting on, this results in a total of 14 LLM labeling runs across the entire dataset.

To combine the LLM-derived labels we choose to ensemble the LLM labels together using majority vote. Specifically, we considered those labels where more than half (i.e., 4 or more of the LLM labelers) agreed on a stance for a comment as producing a stance, otherwise we assigned the label from the LLMs as ``abstain." Although this minorly reduced some of the coverage of the LLM labels, it also improved the quality of the LLM labels overall.

\subsubsection{Use of Analogous Stances} Finally, in the case of labeling the stance of the comment toward its respective video, we also used an analogous stance for a weak label; we used the stance of the comment toward the U.S. Army as a weak label for the stance toward the video. Generally speaking, if a comment was against the U.S. Army, it would also be against a video posted by the U.S. Army, and the use of complementary stances has been shown to improve performance in the stance detection task \cite{sobhani2017dataset, zhang2023investigating}. Thus, we used the final stance toward the U.S. Army produced by weak supervision on the aforementioned weak labels as a weak label for the stance toward the video. It is important to note that this relationship does not work the other way; we found several examples where commentators would express a supportive stance toward the U.S. Army but would be very against a particular video. 

\subsection{Final Stance Determination}
Having created the weak labels for the comments, we then use the weak supervision methodology in the form of the snorkel Python package\footnote{\url{https://snorkel.readthedocs.io/en/master/index.html}}, which uses a unified generative model that takes into account partial coverage of the weak labels as well as correlations between the weak labels to produce the final labels. We refer the reader to \cite{ratner2016data} and \cite{ratner2019training} for the specifics on the function, how it is trained, and how it produces the final labels. For the final labels, we allow the stance to take the form of ``supports", ``against", and ``neutral" for both the stance toward the video and the stance toward the US Army.

\subsection{Human evaluation}
In order to assess the quality of the machine-produced labels, we needed to validate these labels by humans. Since it is impractical to label the entire dataset by humans, we randomly selected 1,000 comments which were labeled by four annotators. We then assessed the label agreement among the annotators and combined the human-labels by majority vote. Using these labels, we assessed the quality of the machine-produced labels.

\section{Data Properties and Analysis}
This section describes some exploratory data analysis performed on this dataset. The dataset consists of 1082 videos published from 1 June 2010 through 1 April 2024 on the US Army's official YouTube Channel. From these videos, there are a total of 215,509 comments.

\subsection{Dataset Properties}

\begin{figure}[!ht]
    \centering
    \includegraphics[width=0.45\textwidth]{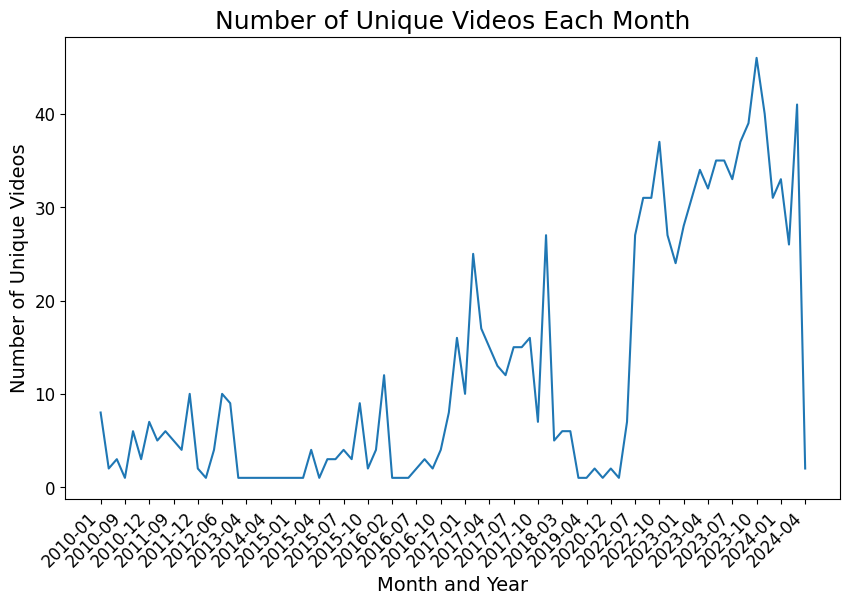}
    \caption{Distribution of videos published over time}
    \label{fig:vid_over_time}
\end{figure}

\paragraph*{Distribution of videos over time}
Generally speaking, the channel has published an increasing number of videos over time, with some occasional periods of fewer videos. \autoref{fig:vid_over_time} depicts the distribution of all the videos on the channel over time. \autoref{tab:per_video_comment_statistics} summarizes the statistics for the comments to each video. Most videos within the data set only have a single comment, with the number of comments per video being highly skewed and some videos having a large number of comments (the most commented-on video, `Soldier wakes up and chooses ``HOOAH!"', has 9.4\% of all of the comments by itself). When it comes to the timing of the comments themselves, we find the overwhelming have occurred recently in the channel's timeline (since 2022).

\begin{figure}[!ht]
    \centering
    \includegraphics[width=0.45\textwidth]{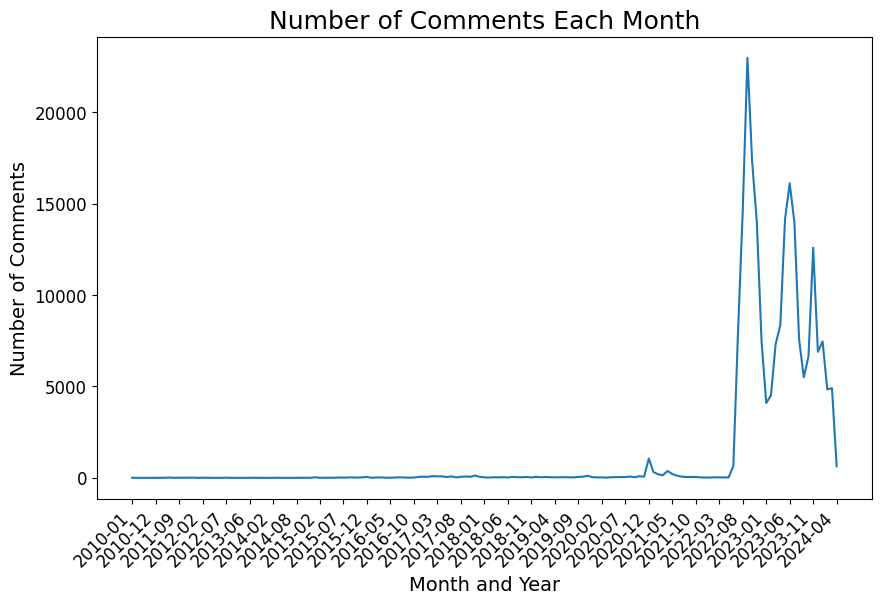}
    \caption{Distribution of comments over time}
    \label{fig:army_video_comments_over_time}
\end{figure}

From \autoref{fig:army_video_comments_over_time}, the plot of comments over time, there are significant spikes in comments in August, September, and October of 2022 and again in May and June of 2023. In the case of the first spike, the most commented-on video was `Soldier wakes up and chooses ``HOOAH!"', while the second spike in the Spring of 2023 has `Words of wisdom from a \#WWII veteran' the most commented-on video. For the first video, it seems the commentary is divisively split about what the Army does or should do in the world, while in the second case the comments seem to both positively reminisce about the Army in World War II and lament about its current state. Turning to the comments themselves, we find that most comments are relatively short, like many social media posts, but a few comments are much longer at over 1,000 words for the comment. \autoref{tab:word_level_comment_statistics} has the word count statistics for the comments.

\begin{table}[ht]
\centering
\begin{tabular}{l|r}
\hline
\textbf{Word-level Statistics of Comments} & \textbf{Value} \\
\hline
Mean & 16.66 \\
Standard Deviation & 33.13 \\
Minimum & 1.00 \\
25th Percentile & 4.00 \\
50th Percentile (Median) & 9.00 \\
75th Percentile & 18.00 \\
Maximum & 1851.00 \\
\hline
\end{tabular}
\caption{Word-level Statistics of the Comments}
\label{tab:word_level_comment_statistics}
\end{table}

\begin{table}[ht]
\centering
\begin{tabular}{l|r}
\hline
\textbf{Statistics of comments per video} & \textbf{Value} \\
\hline
Mean & 199.17 \\
Standard Deviation & 883.41 \\
Minimum & 1 \\
25th Percentile & 7 \\
50th Percentile (Median)& 37 \\
75th Percentile & 86 \\
Maximum & 18999 \\
Mode & 1 \\
\hline
\end{tabular}
\caption{Per Video Comment Statistics}
\label{tab:per_video_comment_statistics}
\end{table}

\subsection{Stance Label Properties}
Turning now to stance labels produced through our methodology for the dataset, we find that the labels skew slightly toward the ``against'' characterization for both the US Army and the videos being posted on the channel. \autoref{tab:stance_counts} summarizes the label counts for each of the stances. 

From the labeling we can see some label skew. First, we observe a label skew toward the ``neutral" label class, which is characteristic of all other stance-labeled datasets (e.g., \citet{covidlies} and \citet{phemerumours}). We do also, however, observe a slight label skew toward ``against" stances versus ``supportive" stances. Since these are comments meant to advocate for a brand (i.e., the U.S. Army), the skew in valenced stances against the U.S. Army and its videos is a cause for concern. Finally, we also observe that through the use of the weak labels, we were unable to label to a couple of thousand data points (around $1\%$ of the data) for each stance.

\begin{table}[ht]
\centering
\begin{tabular}{l|c|c}
\hline
\textbf{Stance} & \textbf{U.S. Army} & \textbf{Video} \\
\hline
abstain  & 2373 (1.10\%) & 1564 (0.73\%) \\
against  & 49817 (23.12\%) & 63222 (29.34\%) \\
neutral  & 122028 (56.61\%) & 102479 (47.56\%) \\
supports & 41291 (19.16\%) & 48244 (22.37\%) \\
\hline
\end{tabular}
\caption{Labels Counts (and percentages of the total number) for Each of the Stance Targets}
\label{tab:stance_counts}
\end{table}

\begin{figure}[!ht]
    \centering
    \includegraphics[width=0.45\textwidth]{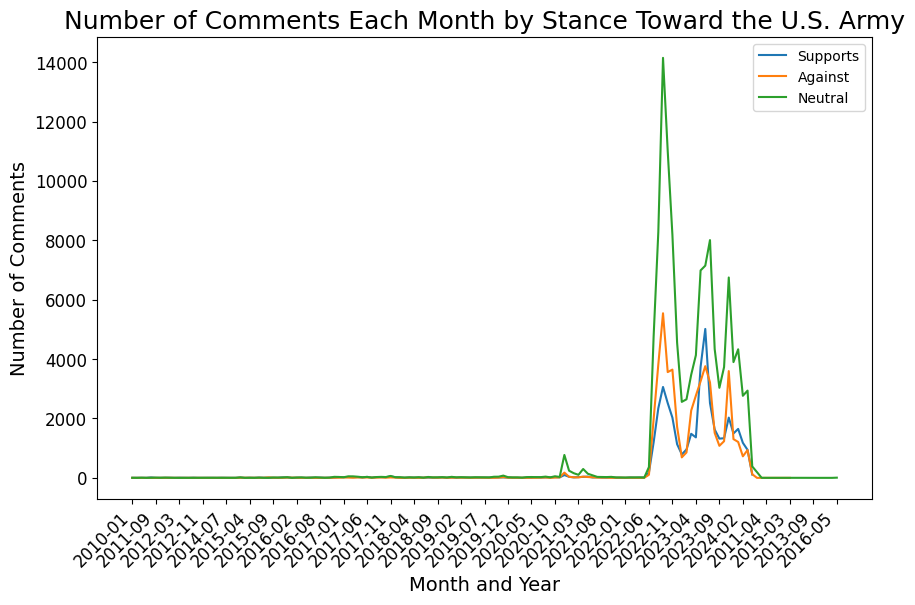}
    \caption{Distribution of videos published on the U.S. Army's Official Channel over time}
    \label{fig:army_video_comments_by_stance_over_time}
\end{figure}

Turning to the distribution of stance labels over time, we find that each of the three classes of stance generally trend together over time. \autoref{fig:army_video_comments_by_stance_over_time} displays the distribution of the different classes of stance toward the US Army for all of the comments. The different stances toward the videos follow a very similar trend. We further observe that the two prominent periods of comments do differ somewhat in their compositions of stance labels. In the first peak of comments, centering in September of 2022, we see more ``against" stance labels than ``supports". Whereas in the second spike of comments around May-June of 2023, we see more ``supports" comments than ``against". Thus, overall, there is a moderate balance among the stances expressed in the comments toward the videos, but this balance does shift over time and between videos. 

To better understand these differences in stances, we looked at the videos with the most positive, neutral and negative receptions. Among the videos with more than 100 comments, we found the video, ``Honoring Our Fallen Heroes - Taps" had the highest percentage of supportive comments at 61.5\% of the comments supporting the video. On the opposite end was the video ``Sergeant Major of the Army Grinston addresses the report of the Fort Hood Independent Review'' which features a short interview with former Sergeant Major Grinston on the investigation at fort Hood following the murder of the Venessa Guillen. The video had 55.9\% of comments that were against it. And, the most neutrally received video was ``Army \#PopQuiz - What training is this?" which received 81.6\% of its comments as being neutral toward the video, many of them answering the question posed in the video.

\begin{figure}[!ht]
    \centering
    \includegraphics[width=0.45\textwidth]{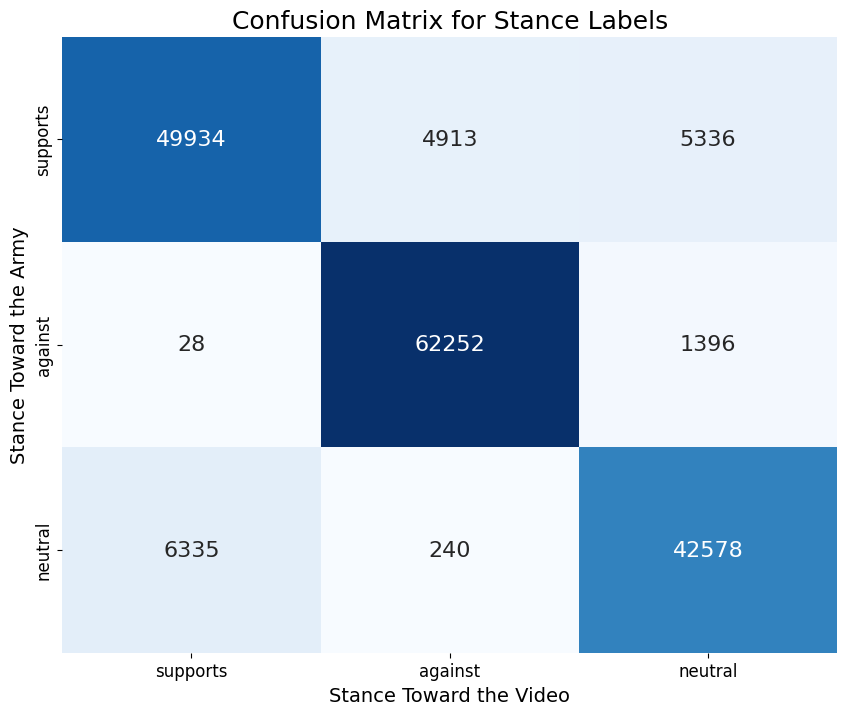}
    \caption{Confusion Matrix of Stance Toward the U.S. Army and Stance Toward the Video}
    \label{fig:stance_labels_confusion_matrix}
\end{figure}

Finally, since this dataset was labeled with more than one stance, we can also look at the relationships between the two stances. \autoref{fig:stance_labels_confusion_matrix}, displays a confusion matrix between the stance toward the U.S. Army and the stance toward the videos. Not unsurprisingly, we find these stances closely resemble each other. It is, however, worth pointing out that we do see a significant number of comments that are supportive of the video, but neutral toward the Army as well as against or neutral to the video, but supportive of the Army. In the first case, from some manual inspection, these are often comments that give a positive word about the video and nothing else. For example, a comment like ``that video was awesome!" fits into this category. For, the second category these comments are often comments that express some frustration or anger at a particular video, but are otherwise supportive of the Army. For example, a comment like ``What is this bs? The Army is better than this..." Overall, the two stance labels closely resemble each other in this dataset, but there are some notable exceptions where they differ, which highlights some of the complexity of stance within this domain and more generally.

\subsection{Comparison with human evaluation}
Finally, we analyzed how well machine labeling, using state--of--the--art techniques, of stance, performed. To do this, we created a validation dataset as previously described and had four human labelers label that data set. For the stance toward the U.S. Army, we had an average Cohen's Kappa \cite{mchugh2012interrater} across all of the human labelers of $0.863$ and for the stance toward the videos of $0.797$. Thus, we were able to obtain high agreement between the labelers on stance of the comments (note that we also released this labeled validation dataset). 

Compared to the output from the weak supervision process, we obtained moderately successful machine labels. For the stance toward the U.S. Army of the comments, the accuracy and Micro F1-scores were $0.68$ and for the stance toward the videos of the comments, the accuracy and Micro F1-scores were $0.639$. When we similarly compared the results of each of the LLMs in the labeling task (see Table \ref{tab:micro_f1_scores}), we do see an improvement of the weak supervision-produced labels over the best-performing LLMs labels, especially for the stance toward the videos.

\begin{table}[ht]
\centering
\begin{tabular}{l|c|c}
\hline
\textbf{Labeling Model} & \textbf{U.S. Army} & \textbf{Video} \\
\hline
mistral zero-shot with comments & 66.90\% & 56.50\% \\
llama-3 zero-shot with comments & 41.40\% & 50.90\% \\
mistral zero-shot & 64.40\% & 55.10\% \\
llama-3 zero-shot & 41.20\% & 40.20\% \\
ul2 zero-shot with comments & 53.00\% & 45.00\% \\
ul2 zero-shot & 49.00\% & 45.80\% \\
mistral adapter & 49.70\% & 26.50\% \\
small model & 52.50\% & 32.30\% \\
\hline
\end{tabular}
\caption{Micro-F1 scores for U.S. Army and Video stances, by different models and prompts, on the validation dataset.}
\label{tab:micro_f1_scores}
\end{table}

From the analysis of the LLM labelers, we notice that there are distinct differences in performance between models and prompting schemes, with the Mistral LLM \cite{jiang2023mistral} and the prompt scheme incorporating replies in the comments generally performing better.

\begin{figure*}[!t]
\centering
\begin{minipage}{0.45\textwidth}
    \centering
    \includegraphics[width=\linewidth]{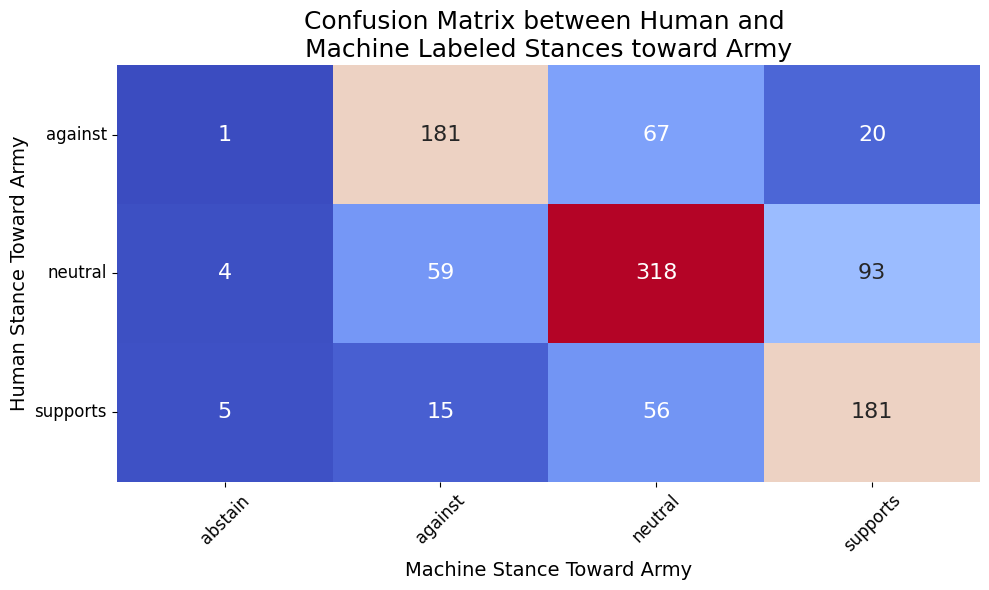}
    \caption{Confusion matrices between the machine-derived labels and the human-derived labels for the validation set.}
    \label{fig:validation_confusion_matrices}
\end{minipage}\hfill
\begin{minipage}{0.45\textwidth}
    \centering
    \includegraphics[width=\linewidth]{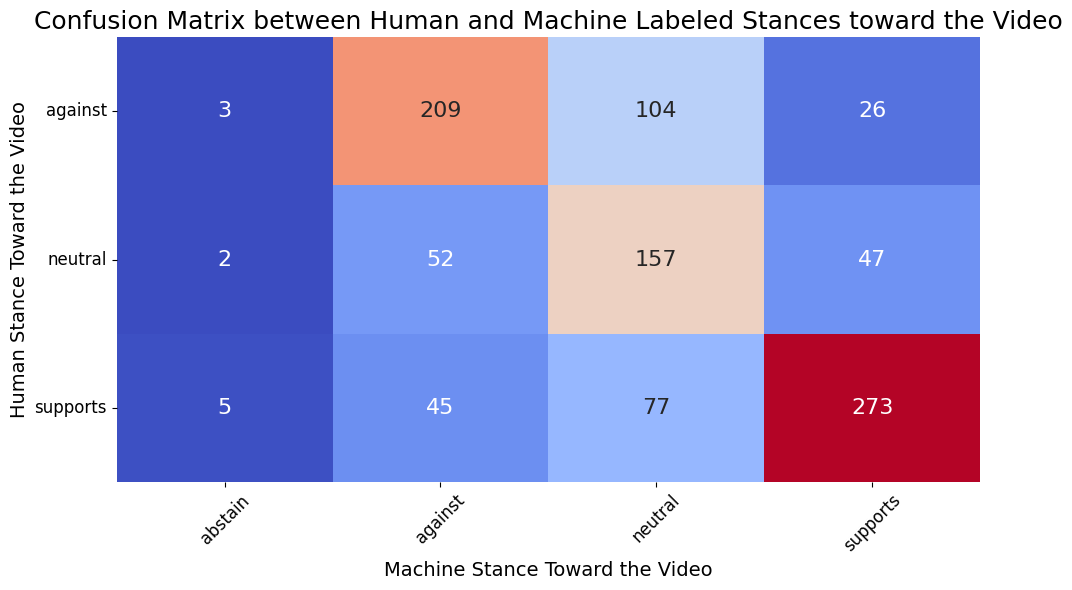}
\end{minipage}
\end{figure*}

To further understand these results, we plotted the confusion matrix between the human-derived and machine-derived labels in Figure \ref{fig:validation_confusion_matrices}. From these matrices, we can observe that for valenced stances, the machine-label stances generally match those from human-annotation; much of loss of accuracy in the machine-labeled stances comes from mis-labeling the neutrally-stanced comments. Upon inspection of some these examples, it seems there are a number of highly-positive sounding comments that are unrelated to the video or the Army as well as some comments where sarcasm or coded language (e.g., `goy', `zog' were missed by the LLMs. 

\section{Discussion} 
Public opinion towards a country's military is crucial, as it signals the level of trust in the government writ large and the willingness of citizens to join the military. A positive opinion facilitates military recruitment and support, especially in times of crisis. Social media engagement plays a pivotal role in shaping this opinion. This paper introduces a dataset that supports such understanding and associated analyses, comprising comments on the US Army's official YouTube videos annotated for their stance towards the Army and the video itself.

To construct this dataset, we employed weak supervision, which leverages \textit{weak labels} for automated labeling. This paradigm provides a transition from direct human annotation to machine-assisted annotation, allowing scalability to larger datasets. For our specific dataset, we utilized subject matter expertise, domain knowledge, and recent advancements in using LLMs for stance determination to create weak labels. Specifically, we used weak labels such as the presence of hate speech, sarcasm, and sentiment as indicators for certain stances. This transformation converted the challenging, ambiguously defined problem of determining and labeling stance into a more well-defined problem for specific aspects or stances. These could then be aggregated through weak supervision to produce comprehensive stance labeling. For instance, the presence of hate speech indicates a comment against the video and, by extension, against the user who uploaded it (e.g., the Army).

We further enhanced these weak labels with contributions from LLMs. Currently, the effectiveness of LLMs in tasks like stance labeling remains uncertain. While recent research shows promise for using LLMs, especially when combined with prompt engineering, for stance detection and labeling, the generalizability of these methods and their compatibility with all LLMs is unclear. Through weak supervision, we treat these LLM-derived labels as noisy yet valuable, enhancing the creation of a stance-labeled dataset. The combination of dataset-specific weak signals and the linguistic understanding power of LLMs produces high-quality labels for stance in a more scalable manner compared to purely manual or machine-generated labels.

In this work we combined the existing state-of-the-art methods for automated labeling of a dataset on the difficult task of stance labeling. We found that these approaches were able to produce reasonable labels, relative to a human-labeled validation dataset, but also that there is still room for improvement in the development of automated labeling techniques, especially for challenging labeling tasks, like stance. This, therefore, opens up avenues for exploration of improving the stance classification methods, especially on out-of-domain tasks, to provide a more accurate label towards an entity and therefore more accurate downstream insights.

\paragraph*{Potential Applications}
The dataset proposed in this work has several potential uses. First, this dataset presents a versatile resource for research on the analysis of stance in YouTube video comments. Beyond its primary application in training and evaluating new stance detection techniques and models, the dataset introduces distinctive traits not present in current benchmarks. Notably, it originates from YouTube, offering a departure from the prevalent X or Twitter-based datasets. Furthermore, the dataset features multiple stance targets and exhibits a substantial degree of stance-taking behavior and controversial content (i.e., a high proportion of ``supports" and ``against" stances). The dataset provides sequential information about comments, allowing investigations into the dynamics of stance with respect to the stance expressed in a previous comment.

Given the dataset's proximity to a politically and culturally divisive topic --- the military --- it encompasses a spectrum of online behaviors, including the propagation of conspiracy theories, trolling, hate speech, and misinformation. As such, it serves as a valuable resource for studies on misinformation and disinformation.

Finally, the dataset's temporal collection allows researchers to explore dynamic aspects of stance-taking behavior over time. This temporal dimension provides insights into the evolution of discussions and sentiments within the context of the US Army's YouTube channel. In summary, the dataset offers a rich foundation for various studies, ranging from advancing stance detection methodologies to delving into the nuanced dynamics of online interactions related to a prominent and controversial subject.

\paragraph*{Dataset Availability}
We collected data provided by the YouTube API and did not perform additional scraping. We only collected public videos and comments, and made no attempt to collect private information. We anonymized the author IDs and did not perform any author-level analysis. The code used for construction of the dataset and the dataset itself is provided at \url{https://doi.org/10.5281/zenodo.10493803}.

\paragraph*{Limitations and Future Work}
Our dataset is limited to the comments from one channel, and the comments we could collect (not all videos posted to the channel allowed comments). We hope to expand our dataset to include the comments towards videos posted by armies around the world, to evaluate public opinion towards the army, across the world. In addition, our dataset is limited to relatively small LLMs due to computational constraints. In the future, we hope to expand the dataset annotation to different types of LLMs with varying prompting schemes to better understand the stance annotation capabilities of LLMs. 

\section{Conclusion}
Understanding the stance expressed in YouTube comments towards an entity is essential for gaining insights into public opinion regarding the authors of the videos. Here, we introduce the DIVERSE dataset, a dataset that captures the stances towards the US Army. This dataset is formulated from comments gathered from the US Army's YouTube videos. We constructed the final stance variable utilizing weak labels, including the identification of hate speech and sarcasm, and incorporated stance inferences from Large Language Models. By employing weak supervision, we amalgamated these diverse labels to generate the final stance annotations. We anticipate that this dataset will prove invaluable to researchers investigating public sentiments and stances toward military entities.

\section*{Acknowledgements}
The research for this paper was supported in part by the Center for Informed Democracy and Social-cybersecurity (IDeaS) at Carnegie Mellon University. This work was also conducted within the Cognitive Security Research Lab at the Army Cyber Institute. The views and conclusions are those of the authors and should not be interpreted as representing the official policies, either expressed or implied, of the Department of Defense, the U.S. Army, or the U.S. Government.

\bibliographystyle{IEEEtranN}
\bibliography{bibliography}

\end{document}